\RequirePackage{snapshot}
\documentclass[letterpaper, 10 pt, conference]{ieeeconf}  
\IEEEoverridecommandlockouts                              

\usepackage{amsfonts,amsmath,amssymb,mathtools}
\usepackage{bigints}
\usepackage{hhline}
\usepackage{marvosym} 

\usepackage{balance}

\renewcommand{\thepage}{}

\usepackage{algorithm}
\usepackage{algpseudocode}
\makeatletter
\let\OldStatex\Statex
\renewcommand{\Statex}[1][0]{%
  \setlength\@tempdima{\algorithmicindent}%
  \OldStatex\hskip\dimexpr#1\@tempdima\relax}
\makeatother

\renewcommand{\algorithmicrequire}{\textbf{Input:}}
\renewcommand{\algorithmicensure}{\textbf{Output:}}
\algnewcommand\AND{~\textbf{and}~}
\algnewcommand\OR{~\textbf{or}~}
\algnewcommand\CONTINUE{~\textbf{continue}~}

\makeatletter
\algnewcommand{\LineComment}[1]{\Statex \hskip\ALG@thistlm \(\triangleright\)
  #1}
\makeatother

\usepackage{graphicx}
\usepackage{booktabs, multicol, multirow}
\usepackage{indentfirst}
\usepackage{subfigure}
\usepackage[font=footnotesize]{caption}
\usepackage[percent]{overpic}

\usepackage{marginnote}
\usepackage[usenames,dvipsnames]{color}
\usepackage[table]{xcolor}    
\definecolor{fullred}{rgb}{0.85,.0,.1} 
\definecolor{navyblue}{rgb}{.0,.0,.5}
\definecolor{bleudefrance}{rgb}{0.19, 0.55, 0.91}
\definecolor{bluegray}{rgb}{0.18, 0.36, 0.6}
\definecolor{lightgray}{rgb}{0.4, 0.4, 0.4}
\definecolor{white}{rgb}{0.9, 0.9, 0.9}

\newcommand{\codecomment}[1]{{\color{lightgray}{#1}}}

\usepackage[textsize=tiny]{todonotes}

\usepackage{array}
\newcolumntype{+}{>{\global\let\currentrowstyle\relax}}
\newcolumntype{^}{>{\currentrowstyle}}

\newcolumntype{C}[1]{>{\centering\arraybackslash}p{#1}}

\makeatletter
\let\NAT@parse\undefined
\makeatother

\usepackage{url}
\usepackage[pagebackref=true,bookmarks=true, colorlinks=true, urlcolor=Blue]{hyperref}

\usepackage{multicol}

\newcommand{\ME}{Sudeep Pillai}
\newcommand{\PAPERAUTHORS}{Sudeep Pillai$^{1}$, Srikumar Ramalingam$^{2}$ and John J. Leonard$^{1}$\\\url{http://people.csail.mit.edu/spillai/fast-stereo-reconstruction}\\\vspace{-5.5ex}}
\newcommand{\PAPERTITLE}{High-Performance and Tunable Stereo Reconstruction}
\newcommand{\PAPERTITLEFMT}{High-Performance and Tunable Stereo Reconstruction\vspace{-2ex}}
\newcommand{\PAPERKEYWORDS}{Stereopsis; Computer Vision; Perception; Obstacle Avoidance}

\hypersetup{pdfauthor={\ME},%
            pdftitle={\PAPERTITLE},%
            pdfsubject={\PAPERTITLE},%
            pdfkeywords={\PAPERKEYWORDS},%
            pdfproducer={LaTeX},%
            pdfcreator={pdfLaTeX}
}

\begin{document}

\title{\PAPERTITLEFMT{}}

\author{
\authorblockN{\PAPERAUTHORS{}}
\thanks{$^{1}$Sudeep Pillai and John J. Leonard are with the Computer
Science and Artificial Intelligence Lab (CSAIL), Massachusetts
Institute of Technology (MIT), Cambridge MA 02139, USA {\tt{\{\href{mailto:spillai@csail.mit.edu}{spillai},
\href{mailto:jleonard@csail.mit.edu}{jleonard}\}\textmd{@csail.mit.edu}}}. Their
work was partially supported by the Office of Naval Research under
grants N00014-10-1-0936, N00014-11-1-0688 and N00014-13-1-0588 and by
the National Science Foundation under grant IIS-1318392.
}
\thanks{$^{2}$Srikumar Ramalingam is with Mitsubishi Electric Research Labs
(MERL)  {\tt{\href{mailto:ramalingam@merl.com}{\textmd{ramalingam@merl.com}}}}, and his work was supported by Mitsubishi
Electric Corporation.}} %



\maketitle

\begin{abstract}
Traditional stereo algorithms have focused their efforts on
reconstruction quality and have largely avoided prioritizing for run
time performance. Robots, on the other hand, require quick
maneuverability and effective computation to observe its immediate
environment and perform tasks within it. In this work, we propose a
high-performance and tunable stereo disparity estimation method, with
a peak frame-rate of 120Hz (VGA resolution, on a single CPU-thread),
that can potentially enable robots to quickly reconstruct their
immediate surroundings and maneuver at high-speeds. Our key
contribution is a disparity estimation algorithm that iteratively
approximates the scene depth via a piece-wise planar mesh from stereo
imagery, with a fast depth validation step for semi-dense
reconstruction. The mesh is initially seeded with sparsely matched
keypoints, and is recursively tessellated and refined as needed (via a
resampling stage), to provide the desired stereo disparity
accuracy. The inherent simplicity and speed of our approach, with the
ability to tune it to a desired reconstruction quality and runtime
performance makes it a compelling solution for applications in
high-speed vehicles.

\end{abstract}

\IEEEpeerreviewmaketitle


\section{Introduction}
\label{sec-introduction}



Stereo disparity estimation has been a classical and well-studied
problem in computer vision, with applications in several domains
including large-scale 3D reconstruction, scene estimation and obstacle
avoidance for autonomous driving and flight etc. Most state-of-the-art
methods~\cite{zbontar2015computing} have focused its efforts on
improving the reconstruction quality on specific
datasets~\cite{scharstein2002taxonomy, Geiger2012CVPR}, with the
obvious trade-off of employing sophisticated and computationally
expensive techniques to achieve such results. Some recent methods,
including Semi-Global Matching~\cite{hirschmuller2005accurate}, and
ELAS~\cite{geiger2011efficient}, have recognized the necessity for
practical stereo matching applications and their real-time
requirements. However, none of the state-of-the-art stereo methods
today can provide meaningful scene reconstructions in real-time
($\geq25$Hz) except for a few FPGA or parallel-processor-based
methods~\cite{honegger2014real,banz2010real,gehrig2009real}. Other
methods have achieved high-speed performance by matching fixed
disparities, fusing these measurements in a push-broom fashion with a
strongly-coupled state estimator~\cite{barry2015pushbroom}. Most
robotics applications, on the other hand, require real-time
performance guarantees in order for the robots to make quick decisions
and maneuver their immediate environment in an agile
fashion. Additionally, as requirements for scene reconstruction vary
across robotics applications, existing methods cannot be reconfigured
to various accuracy-speed operating regimes.



%
\begin{figure}[!t]
  \begin{overpic}[width=\columnwidth]{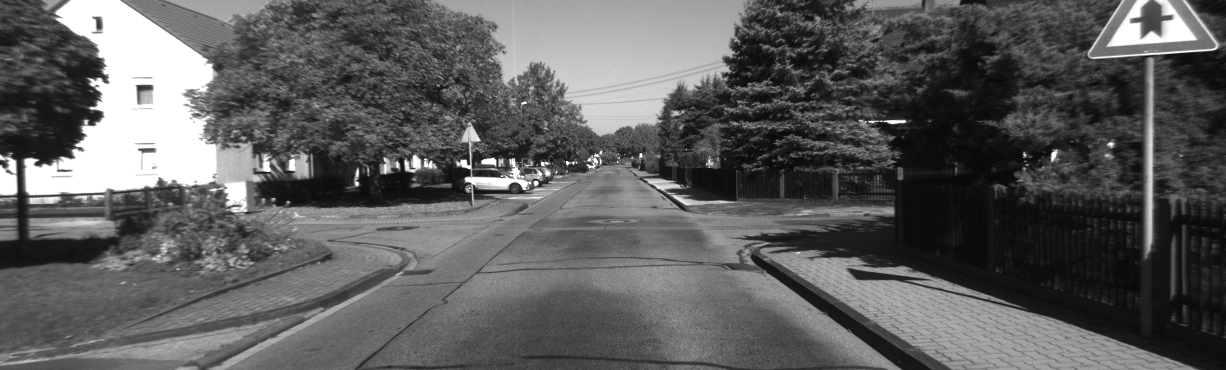} \put(97,1){\footnotesize\color{white}{\textbf{A}}}\end{overpic}
  \begin{overpic}[width=\columnwidth]{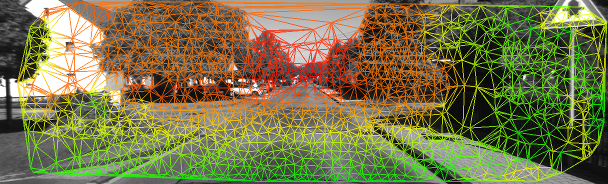} \put(97,1){\footnotesize\color{white}{\textbf{B}}}\end{overpic}
  \begin{overpic}[width=0.5\columnwidth]{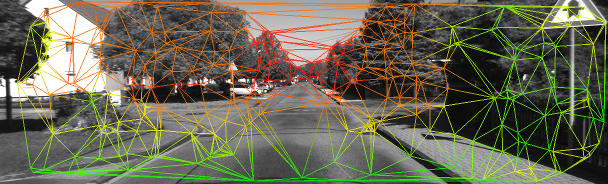} \put(93,2){\footnotesize\color{white}{\textbf{C}}}\end{overpic}\begin{overpic}[width=0.5\columnwidth]{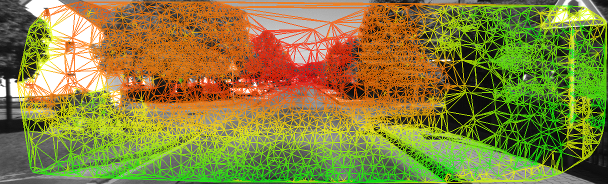} \put(93,2){\footnotesize\color{white}{\textbf{D}}}\end{overpic}
  \begin{overpic}[width=\columnwidth]{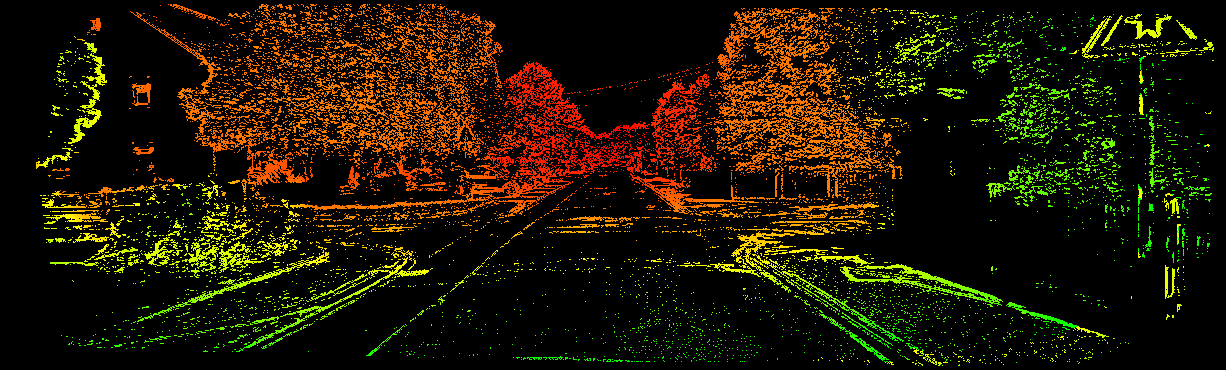} \put(97,1){\footnotesize\color{white}{\textbf{E}}}\end{overpic}
    
  \caption{The proposed high-performance stereo matching method provides semi-dense
    reconstruction (\textbf{E}) of the scene, capable of running at a peak frame-rate of ~120Hz
    (8.2 ms, VGA resolution). Our approach maintains a piece-wise planar
    representation that enables the computation of disparities (semi-densely, and
    densely) for varied spatial densities over several iterations (\textbf{B}-2 iterations, \textbf{C}-1
    iteration, \textbf{D}-4 iterations). Colors illustrate the scene
    depths, with green indicating near-field and red indicating far-field
    regions. Figure best viewed in digital format.}
  \label{fig:intro-fig}
  \end{figure}

In this work, we propose a high-performance, iterative stereo matching
algorithm, capable of providing semi-dense disparities at a peak
frame-rate of 120Hz (see Figure~\ref{fig:intro-fig}). An iterative
stereo disparity hypothesis and refinement strategy is proposed that
provides a tunable iteration parameter to adjust the
accuracy-versus-speed trade-off requirement on-the-fly. Through
experiments, we show the strong reliability of disparity estimates
provided by our system despite the low computational requirements. We
provide several evaluation results comparing accuracies against
current stereo methods, and provide performance analysis for varied
runtime requirements. We validate the performance of our system on
both publicly available datasets, and commercially available stereo
sensors for comparison. In addition to single view disparity
estimates, we show qualitative results of large-scale stereo
reconstructions registered via stereo visual odometry, illustrating
the consistent stereo disparities our approach provides on a per-frame
basis.

\newpage
\section{Related Work}
\label{sec:related-work}

Classical stereo matching methods have mostly considered dense reconstructions,
and are generally divided into two categories, \textit{local} and
\textit{global} methods. The naive approach to stereo matching involves finding
corresponding pixels in the left and right images that have similar color or
intensity. Since the intrinsics and extrinsics of the stereo cameras are known,
the matching search space is limited to the epipolar line with a pre-defined
disparity level, assuming a maximum distance observed. 


\vspace{2mm}
\textbf{Dense Methods} 
As one may expect, the above formulation results in a noisy disparity map, due
to the high pixel-level ambiguity in matching. This is addressed by matching
fixed size windows instead, reducing the noise and inherent ambiguity in
the stereo imagery. Additionally, the resulting disparity is smoothed, allowing
neighboring pixels to have similar disparities. Despite several advances in
adaptive-supports, slanted window matching and edge-preserving filtering
approaches~\cite{bleyer2013stereo}, local methods suffer from being unable to
estimate disparities at low-textured regions.

For the past decade, global methods have dominated stereo
benchmarks~\cite{scharstein2002taxonomy,Geiger2012CVPR}. They differ from local
methods in that their smoothness regularization assumptions are no longer
limited to a fixed window size, but extend throughout the image. Typically, the
disparity estimation is modeled as an energy minimization, given by:
\begin{align} 
E(D) = \sum_{p\in I_l} c(p, p-d_p) + \lambda \sum_{\{p,q\} \in
\mathcal{N}} s(d_p, d_q)
\end{align} where $c(p, p-d_p)$ is the pixel matching cost for a disparity level
$d_p$, $s(d_p, d_q)$ is the smoothness regularization or penalty enforced
between pixels $p$ and $q$ that are neighbors defined by $\mathcal{N}$. The
above energy minimization formulation allows several optimization strategies to
be employed including (i) graph-cuts (ii) belief-propagation (iii) dynamic
programming. For a more thorough description of state-of-the-art stereo
matching, we refer the reader to~\cite{bleyer2013stereo}.


\vspace{2mm}
\textbf{Sparse and Semi-Dense Methods} Sparse stereo matching methods have been
prevalent in robotics applications primarily due to their low-computational
complexity~\cite{schauwecker2012new}. These methods, including monocular
keypoint-based SLAM techniques, have been combined with tessellation or meshing
techniques to represent the scene as piece-wise planar~\cite{concha2015}, making
it a fairly rich representation for navigation and scene reconstruction purposes
with a significantly low memory footprint.


Recently, there has been an increased interest in semi-dense
representations for mapping,
navigation~\cite{veksler2002dense,engel2014lsd,Mur-Artal-RSS-15,ramalingam2015line}
and object detection~\cite{Pillai-RSS-15}. Qualitatively, these
semi-dense methods can be a compelling middle-ground, between dense
stereo and sparse stereo matching methods, potentially paving the way
to newer representations for navigation and
reconstruction. LSD-SLAM~\cite{engel2014lsd}, has recently shown
large-scale 3D reconstructions by fusing the depth estimates for
high-gradient pixels from short and wide-baseline frames in monocular
videos, without the use of any interest point matches. However,
monocular methods suffer from the well-known scale-drift problem
(corrected using an IMU),
and rely on the availability of several images to provide
metrically accurate reconstructions. Recently, a semi-dense stereo
reconstruction of high gradient pixels was shown using a Line-Sweep
algorithm~\cite{ramalingam2015line}, which uses cross-ratio
constraints on locally planar region. Our method relies on Delaunay
triangulation and support point re-sampling, leading to better
accuracy and improved computational performance. Furthermore, our
method can reconstruct heavily occluding objects like poles, which
will be challenging for Line-sweep.


\vspace{2mm}
\textbf{Depth-priors and Plane-based Stereo} Our work closely relates
to that of ELAS~\cite{geiger2011efficient} that takes a generative
approach, using tessellated support points from sparse stereo matching
as a depth prior to enable efficient sampling of disparities in a
dense fashion. Most recently, MeshStereo~\cite{zhang2015meshstereo}
has been proposed, where the global stereo model is designed for view
interpolation via a similar 3D triangular mesh. The authors model the
difficult depth discontinuity problem as a two-layer MRF, where the
upper layer models the splitting of depth discontinuities, while the
lower layer regularizes the depths via a region-based optimization. In
this work, we take a discriminative approach to stereo matching, and
continue to maintain the piece-wise planar assumption while
re-tessellating poorly reconstructed regions in the interpolated
disparity image that correspond to having a high matching
cost. Furthermore, we propose an iterative method that continues to
re-tessellate and approximate complex surfaces with more piece-wise
planar regions, with every additional iteration.

Similar to Patch-Match Stereo~\cite{bleyer2011patchmatch}, our method
implicitly computes disparities with sub-pixel precision, without the
need for an additional post-processing step~\cite{yang2007spatial}
that fits a parabolic curve within the cost volume. As duly noted
in~\cite{bleyer2013stereo}, parabolic fitting leads to noisy sub-pixel
estimation across heavily slanted surfaces. We do note that our
approach is reminiscent of plane-sweeping algorithms that include
fronto-parallel and slanted windows to their label space for improved
disparity estimation along varied surfaces~\cite{gallup2007real},
however, we draw candidate planes and disparities from the
tessellations constructed with sparse keypoint-based stereo matches
that in turn reduces the search space drastically.



\vspace{2mm}
\textbf{High-speed Stereo Matching} To the best of our knowledge, we
are unaware of any semi-dense stereo method that can perform full
disparity range estimation at speeds of $\geq 100$Hz, without the use
of GPUs, FPGAs or other specialized-hardware. We consider disparity
estimation for the approximate piece-wise planar case, as this
representation can be especially useful in robotics applications where
obstacles are to be observed and avoided in real-time.  We propose an
iterative stereo matching method, that maintains a spatially-adaptive
piece-wise planar representation, significantly speeding up stereo
disparity estimation by a factor of 32x compared to popular stereo
implementations~\cite{hirschmuller2005accurate}, while providing
sufficiently accurate disparity estimates.

\newpage
\section{High-Performance and Tunable Stereo Reconstruction}
\label{sec:proc-procedure}

This section introduces the algorithmic components of our method~(see
Alg.~\ref{alg:main-algorithm}). We propose a tunable (and iterative)
stereo algorithm that consists of four key steps: (i)
Depth prior construction from Delaunay triangulation of sparse
key-point stereo matches (ii) Disparity interpolation using piece-wise
planar constraint imposed by the tessellation with known depths (iii)
Cost evaluation step that validates interpolated disparities based on
matching cost threshold (iv) Re-sampling stage that establishes new
support points from previously validated regions and via dense
epipolar search. The newly added support points are re-tessellated and
interpolated to hypothesize new candidate planes in an iterative
process. Since we are particularly interested in collision-prone
obstacles and map structure in the immediate environment, we focus on
estimating the piece-wise planar reconstruction as an approximation to
the scene, and infer stereo disparities in a semi-dense fashion from
this underlying representation. Unless otherwise noted, we consider and
perform all operations on only a subset of image pixels that have high
image gradients $\Omega_{I} \subset \Omega$, and avoid reconstructing
non-textured regions in this work.

%
\begin{figure}[!b]
  \centering
  {\renewcommand{\arraystretch}{0.4} 
    {\setlength{\tabcolsep}{0.1mm}
      \vspace{3mm}
    \begin{tabular}{c}
    \includegraphics[width=0.98\columnwidth]{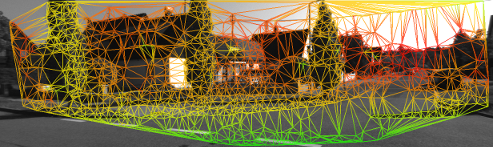}\\
    \includegraphics[width=0.98\columnwidth]{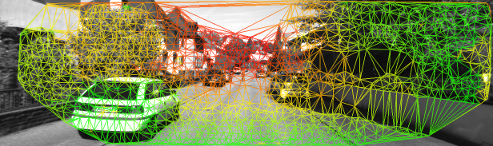}
  \end{tabular}}}
  \caption{Depth prior determined via Delaunay triangulation of sparse support
    points. Vertices in the mesh correspond to the sparse support points, or
    re-sampled support, while the triangular regions represent the piece-wise planar
    scene reconstruction. }
  \label{fig:disparity-edge-scenes}
\end{figure}

\subsection{Spatial Support via Sparse Stereo Matching}
\label{sec:proc-procedure-sparse-stereo} Many state-of-the-art stereo algorithms
start by exhaustively computing a pixel-level cost volume
$\mathcal{O}(HWN_{D})$, for a fixed number of disparities $N_{D}$ (usually
128). Instead, we employ a similar strategy to~\cite{geiger2011efficient}, and
first construct a piece-wise planar scene depth estimate to quickly inform a
coarse depth prior or mesh. First, a sparse set of support keypoints
$S=\{s_1,\dots,s_n\}$ are detected via FAST
features~\cite{rosten2006machine} (sampled from 12x10 spatial-bins), and matched along their epipolar
lines as in~\cite{schauwecker2012new}~(see~\textproc{\footnotesize SparseStereo} in
Alg.~\ref{alg:main-algorithm}).  We define each support point $s_n =
(u_n,v_n,d_n)^T$, similar to~\cite{geiger2011efficient}, as the concatenation of
their image coordinates $(u_n,v_n) \in \mathbb{N}^2$, and their corresponding
disparity $d_n \in \mathbb{N}$. Using these support points as vertices with
known depths, a piece-wise planar mesh is constructed via 
Delaunay-Triangulation. (see Figure~\ref{fig:disparity-edge-scenes},
~\textproc{\footnotesize DelaunayTriangulation} in
Alg.~\ref{alg:main-algorithm}).


%
%
{
  \begin{algorithm}[!t]
    \footnotesize
    \caption{\small Iterative Stereo Reconstruction}
    \label{alg:main-algorithm}
    \begin{algorithmic}[1]
      \Require $(I_l,I_r,\Omega_{I})$: Input gray-scale stereo images and
      high-gradient regions
      \Ensure $D_f$: Disparities at high-gradient regions (Semi-Dense)
      \Statex[-1]\hspace{-0.5mm}\textbf{Globals}: Refer to
      Table~\ref{table:nomenclature} for description of variables
      \vspace{2mm}
      
      \Statex[0] \codecomment{// Initialize final disparity and associated cost}
      \State $D_f \gets [0]_{[H \times W]}, C_f \gets [t_{hi}]_{[H \times W]}, sz_{occ} \gets 32$
      
      \vspace{1mm}

      \Statex[0] \codecomment{// $S$: Set of N support
        points}
      \State $S_1 \gets$ \Call{SparseStereo}{$I_l,I_r$} 
      \vspace{1mm}      
      \Statex[0] \codecomment{// Tessellated mesh with estimated disparities}      
      \State $\mathcal{G}(S_{1}) \gets$ \Call{DelaunayTriangulation}{$S_{1}$}

      \vspace{2mm}
      \For{$it=1\rightarrow n_{iters}$}

      \vspace{1mm}       
      \Statex[1] \codecomment{// Dense piece-wise planar disparity}      
      \State $D_{it} \gets$ \Call{DisparityInterpolation}{$\mathcal{G}(S_{it})$}

      \vspace{1mm}
      \Statex[1] \codecomment{// Cost evaluation given interpolated disparity}
      \State $C_{it} \gets$ \Call{CostEvaluation}{$I_l,I_r,D_{it}$}

      \vspace{1mm}
      \Statex[1] \codecomment{// Refine disparities}
      \State $C_g,C_b \gets$ \Call{DisparityRefinement}{$D_{it},C_{it}$}
      
      \vspace{1mm}
      \Statex[1] \codecomment{// Prepare for next iteration, if not last iteration}
      \If {$it\neq n_{iters}$}

      \vspace{1mm}
      \Statex[2] \codecomment{// Re-sample regions with high matching cost}
      \State $S_{it+1} \gets$ \Call{SupportResampling}{$C_g,C_b,S_{it}$}

      \vspace{1mm}
      \Statex[2] \codecomment{// Tessellated mesh with estimated disparities}      
      \State $\mathcal{G}(S_{it+1}) \gets$ \Call{DelaunayTriangulation}{$S_{it+1}$}

      \vspace{1mm}
      \Statex[2] \codecomment{// Decrease occupancy grid size by factor of 2}      
      \State $sz_{occ} = \max(1, sz_{occ} / 2)$
      \vspace{1mm}      
      \EndIf
      
      \EndFor
    \end{algorithmic}
  \end{algorithm}
}

\begin{table}[!t]
\centering
\scriptsize
{
\setlength{\tabcolsep}{0.4em}
\begin{tabular}{lcl}
\textbf{Name} & \textbf{Scope} & \textbf{Description}\\\midrule
$I_l, I_r$  & L & Input gray-scale stereo images\\ 
$H, W$  & G & Dimensions of input image $I_l$\\ 
$\Omega, \Omega_{I}$ & G & Set of all pixels in image, and subset of high-gradient pixels\\ 
$S$   & L & Sparse support pixels with valid depths\\ 
$\mathcal{G}(S)$ & L & Graph resulting from Delaunay Triangulation over $S$ \\ 
$X$   & L & Re-sampled or detected support pixels with unknown depths\\ 
$D_f$   & G & Final disparity image \\ 
$C_f$   & G & Cost matrix associated to $D_f$  \\ 
$D_{it}$   & L & Intermediate disparity (interpolated)   \\ 
$C_{it}$   & L & Cost associated to $D_{it}$ \\ 
$C_g$   & L & Cost associated with regions of high confidence matches \\ 
$C_b$   & L & Cost associated with regions of invalid disparities \\ 
$N_D$   & G & Maximum number of disparities considered \\ 
$sz_{occ}$   & G & Occupancy grid size used for re-sampling  \\ 
$t_{lo}, t_{hi}$   & G & Lower and upper cost threshold for
                     validating disparities \\ 
$n_{iters}$   & G & Number of iterations the algorithm is allowed to run\\ 
\bottomrule %
\end{tabular}\\\vspace{1mm}
}
\caption{Description of symbols used in the proposed stereo matching algorithm,
  and their corresponding scope (G:Global or L:Local) within the implementation.}
\label{table:nomenclature}
\end{table}

\subsection{Disparity Interpolation} We refer to the planar regions in the
delaunay triangulation as candidate planes, as they are constructed from the
sparse set of support points whose disparities are estimated via epipolar
search. These candidate planes provide a strong measure of an underlying
surface, and can be used to quickly verify the hypothesized planes. Inspired by
previous work on candidate-plane validation~\cite{bleyer2011patchmatch}, we
leverage this efficient verification step to iteratively hypothesize candidate
regions in the disparity image, thereby limiting the effective disparity search
space to fewer than 3-5 disparity levels (not limiting to integer-valued
disparities as most dense methods do).

At every intermediate step, we treat the stereo disparity image $D_{it}$ as being constructed in a
piece-wise planar manner via the Delaunay tessellated mesh. Each 3D planar
surface or triangle, can be described by its 3D plane parameters
$(\pi_1,\pi_2,\pi_3,\pi_4) \in \mathbb{R}^4$ given by
\begin{align} \pi_1X + \pi_2Y + \pi_3Z + \pi_4 = 0
\label{eq:3d-plane}
\end{align} For a stereo setup with a known baseline $B$, and known calibration
($u = fX / Z$, $v = fY / Z$, and $d = fB / Z$), the above equation reduces to
\begin{align} \pi_1'u+ \pi_2'v + \pi_3' = d
\label{eq:disparity-plane}
\end{align} where $\pi' = (\pi_1',\pi_2',\pi_3') \in \mathbb{R}^3$ are the plane
parameters in disparity space.

In order to estimate interpolated disparities on a pixel-level basis, we first
construct a lookup-table that identifies the triangle and its plane coefficients
for each pixel $(u,v)$ in the left image. Subsequently, the parameters $\pi'$
for each triangle are obtained by solving a linear system as done
in~\cite{geiger2011efficient}, and are re-estimated every time after the
Delaunay triangulation step. The resulting piece-wise planar tessellation can be
used to linearly interpolate regions within the disparity image using the
estimated plane parameters $\pi'$~(see~\textproc{\footnotesize
DisparityInterpolation} in Alg.~\ref{alg:main-algorithm}).


\subsection{Cost Evaluation} The interpolated disparity image resulting from
every tessellation provides a set of candidate depths that could potentially
contain valid scene points. In order to validate these interpolated disparities,
we perform Census window-based matching on a 5x5 patch~\cite{zabih1994non,
hirschmuller2009evaluation} between the left and right stereo images. The
resulting matching cost is normalized and retained to be validated in the next
step~(see~\textproc{\footnotesize CostEvaluation} in
Alg.~\ref{alg:cost-evaluation}).

%
%
{
  \begin{algorithm}[t]
    \footnotesize
    \caption{\small \textproc{CostEvaluation}}
    \label{alg:cost-evaluation}
    \begin{algorithmic}[1]
      \Require $(I_l, I_r, D_{it})$: Left/Right stereo image, and interpolated disparity
      \Ensure $C_{it}$: Matching cost corresponding to $D_{it}$
      \vspace{2mm}

      \For {$(u,v) \in \Omega_{I}$}

      \vspace{1mm}
      \Statex[1] \codecomment{// Interpolated disparity at $(u,v)$}
      \State $d \gets D_{it}(u,v)$


      \vspace{1mm}
      \Statex[1] \codecomment{// Census-based 5x5 window matching}
      \State $C_{it}(u,v) \gets$ \Call{CensusMatchingCost}{$I_l(u,v), I_r(u-d,v)$}
            
      \EndFor
    \end{algorithmic}
  \end{algorithm}
}

\subsection{Disparity Refinement} The interpolated disparities
computed from the tessellation may or may not necessarily hold true
for all pixels. For high-gradient regions in the image, the cost
computed between the left and right stereo patch for the given
interpolated disparity can be a sufficiently good indication to
validate the candidate pixel disparity. We use this assumption to
further refine and prune candidate disparities based on the per-pixel
cost computed in the previous step, as characterized by validated
($C_g$) and invalidated ($C_b$) cost regions. Thus, we can invalidate
every pixel $p$ in the left image, if the cost associated
$c(p,p-d_{i})$ with matching the pixel in the right image with a given
interpolated disparity $d_{i}$ is above a maximum permissible cost
$t_{hi}$. The same approach is used to validate pixels that fall within
a suitable cost range ($<t_{lo}$) whose correspondence certainty is
high. This step also allows incorrectly matched regions to be
resampled and re-evaluated for new stereo matches as the interpolated
costs of regions around the falsely matched corners are driven
sufficiently high. Additionally, the disparities corresponding to the
least cost for each pixel is updated with every added iteration,
ensuring that the overall stereo matching cost is always reduced~(see
Step~\ref{algref:minimized-cost} in
Alg.~\ref{alg:disparity-refinement}). For more details regarding this
step see~\textproc{\footnotesize DisparityRefinement} in
Alg.~\ref{alg:disparity-refinement}.

%
%
{
  \begin{algorithm}[t]
    \footnotesize
    \caption{\small \textproc{DisparityRefinement}}
    \label{alg:disparity-refinement}
    \begin{algorithmic}[1]
      \Require $(D_{it},C_{it})$: Interpolated Disparity and associated matching
      cost
      \Ensure $C_g, C_b$: Costs associated with regions of high and low matching
      confidence disparities
      \vspace{2mm}

      \State $H' \gets \frac{H}{sz_{occ}}, W' \gets \frac{W}{sz_{occ}}$
      \vspace{1mm}
      \Statex[0] \codecomment{// $C_g$: Cost matrix of confident supports: $(u,v,d,cost)$}
      \State $C_g \gets [0,0,0,t_{lo}]_{[H' \times W']}$
      \vspace{1mm}
      \Statex[0] \codecomment{// $C_b$: Cost matrix of invalid matches: $(u,v,cost)$}
      \State $C_b \gets [0,0,t_{hi}]_{[H' \times W']}$ 

      \vspace{1mm}
      \For {$(u,v) \in \Omega_{I}$}


      \vspace{1mm}
      \Statex[1] \codecomment{// Establish occupancy grid for resampled points}
      \State $u' \gets \frac{u}{sz_{occ}}, v' \gets \frac{v}{sz_{occ}}$

      \vspace{1mm}
      \Statex[1] \codecomment{// If matching cost is lower than
        previous best final cost}
      \If {$C_{it}(u,v) < C_f(u,v)$}\label{algref:minimized-cost}
      \State $D_f(u,v) \gets D_{it}(u,v)$
      \State $C_f(u,v) \gets C_{it}(u,v)$
      \EndIf

      \vspace{1mm}
      \Statex[1] \codecomment{// If matching cost is lower than
        previous best valid cost}
      \If {$C_{it}(u,v) < t_{lo} \AND C_{it}(u,v) < C_g(u',v',4^{\dagger})$}
      \State $C_g(u',v') \gets (u,v,D_{it}(u,v),C_{it}(u,v))$
      \EndIf
      
      \vspace{1mm}
      \Statex[1] \codecomment{// If matching cost is higher than
        previous worst invalid cost}
      \If {$C_{it}(u,v) > t_{hi} \AND C_{it}(u,v) > C_b(u',v',3^{\dagger})$}
        \State $C_b(u',v') \gets (u,v,C_{it}(u,v))$
      \EndIf
      
      \EndFor
    \end{algorithmic}
  \hfill\scriptsize{$^\dagger$Matrices are 1-indexed}
\end{algorithm}
  }
%
%

%
%
{
  \begin{algorithm}[t]
    \footnotesize
    \caption{\small \textproc{SupportResampling}}
    \label{alg:support-resampling}
    \begin{algorithmic}[1]
      \Require $(C_{g},C_{b}, S_{it})$: Matching costs for confident/invalid matches
      \Ensure $S_{it+1}$: New support points for tessellation
      \vspace{2mm}

      \State $S_{it+1} \gets S_{it}, X \gets \emptyset$
      \For {$(u,v) \in \Omega_{I}$}

      \vspace{1mm}
      \Statex[1] \codecomment{// Perform sparse epipolar stereo for resampled
        invalid pixels}
      \If {$C_b(u,v) \neq 0$}
      \State $X \gets \{X, (u,v)\}$
      \EndIf

      \vspace{1mm}
      \Statex[1] \codecomment{// Resample confident pixels and add to support}
      \If {$C_g(u,v) \neq 0$}
      \State $S_{it+1} \gets \{S_{it+1}, (u,v)\}$
      \EndIf
            
      \EndFor

      \vspace{2mm}
      \Statex[0] \codecomment{// Re-estimate disparities via epipolar search}
      \State $S_{matched} \gets$ \Call{SparseEpipolarStereo}{$I_l,I_r,X$}
      \State $S_{it+1} \gets \{S_{it+1}, S_{matched}\}$

    \end{algorithmic}
  \end{algorithm}
}

\subsection{Support Resampling} The disparity refinement step establishes pixels
or regions in the image whose disparities need to be re-evaluated, while also
simultaneously providing reliable disparities to further utilize in the matching
process. With a discretized occupancy grid of size ($sz_{occ} \times sz_{occ}$),
pixels with the highest matching cost within a 32x32 ($sz_{occ}$ is initialized
to 32) window are established, and re-sampled. These re-sampled pixels are
strong indicators of occluding edges, and sharp discontinuities in depth, making
them viable candidates for epipolar-constrained dense stereo
matching. Subsequently, the re-sampled keypoints are densely matched via
epipolar search, and new support points $S_{matched}$ are established as a
result. Another valuable feature is the ability to inform disparities at greater
resolution and accuracy with every subsequent iteration; the discretization of
the occupancy grid is reduced by a factor of 2 so that pixels are more densely
sampled with every successive iteration~(see~\textproc{\footnotesize
SupportResampling} in Alg.~\ref{alg:support-resampling}).

\subsection{Iterative Reconstruction} 
\begin{figure}[h]
  \centering
  {\renewcommand{\arraystretch}{0.2} 
    {\setlength{\tabcolsep}{0.1mm}
    \begin{tabular}{cc}
    \includegraphics[width=0.5\columnwidth]{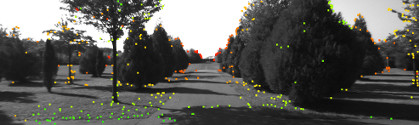}&
    \includegraphics[width=0.5\columnwidth]{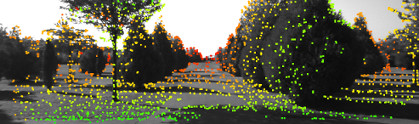}\\
    \includegraphics[width=0.5\columnwidth]{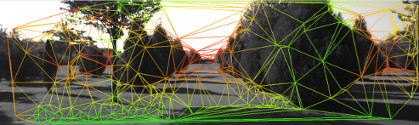}&
    \includegraphics[width=0.5\columnwidth]{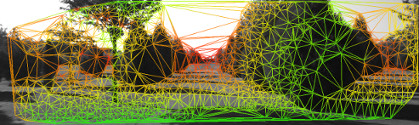}\\
    \multicolumn{2}{c}{\includegraphics[width=\columnwidth]{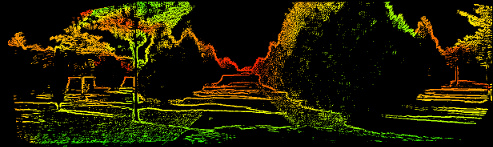}} \\
  \end{tabular}}}
\caption{Depth prior estimated with every subsequent iteration (Rows 1-2, Column
  1: After 1 iteration, Column 2: After 2 iterations). As expected, the density of
  support points increase, with the piece-wise planar representation better
  fitting to the true scene disparity map. Row 3 illustrated the final semi-dense
  reconstruction after 2 iterations. Figure best viewed in digital format.}
  \label{fig:disparity-edge-iterations}
\end{figure}

The stereo matching proceeds to reduce the
overall stereo matching cost associated with the interpolated piece-wise planar
disparity map. High-matching cost regions are re-sampled and re-estimated to
better fit the piece-wise planar disparity map to the true scene
disparity. With every subsequent iteration, new keypoints are sampled,
tessellated to inform a piece-wise planar depth prior, and further evaluated to
reduce the overall matching cost. With such an iterative procedure, the overall
stereo matching cost is reduced, with the obvious cost of added computation or
run-time requirement (see Figure~\ref{fig:disparity-edge-iterations}).

%

\section{Experiments}
\label{sec:experiments}

In this section, we evaluate the proposed high-performance stereo matching
method. We evaluate the matching accuracy and runtime performance of our
proposed method on the popular KITTI dataset~\cite{Geiger2012CVPR} and on 2
different stereo cameras, namely the Point Grey Bumblebee2
1394a\footnote{\scriptsize~\url{http://www.ptgrey.com/stereo-vision-cameras-systems}}, and
the ZED Stereo Camera\footnote{\scriptsize~\url{https://www.stereolabs.com/zed/}}. The
KITTI dataset contains rectified gray-scale stereo imagery at a resolution of
1241x376 (0.46 MP), captured from 2 Point Grey Flea2 cameras mounted with a
baseline of 0.54m. We compare against stereo matching algorithms that are
commonly used in robotics applications - the popular implementation of
Semi-Global Matching~\cite{hirschmuller2005accurate} in OpenCV (Semi-Global
Block-Matching or SGBM), ELAS~\cite{geiger2011efficient} and
Line-Sweep~\cite{ramalingam2015line}. We provide a thorough analysis of the
trade-offs between matching accuracy and run-times achievable by our proposed
method, across varied hardware and environmental setups.

\subsection{Evaluation on KITTI dataset} \textbf{Disparity Estimation
Accuracy} In order to evaluate our proposed semi-dense method against
existing methods, we only consider disparities in the image that have
large image gradients or edges. Currently, semi-dense methods cannot
be fully evaluated on the KITTI dataset, since the test server
interpolates missing disparities, introducing several errors in the
disparity estimates and overall accuracy. For all valid and
non-occluding semi-dense edges, we report the absolute difference
between the proposed method and existing state-of-the-art stereo
implementations. In our experiments on the provided KITTI stereo
evaluation kit, we find that greater than 89.9\% of edge pixels had a
disparity value of less than 3 pixels with respect to ground truth for
the single pass variant (\textit{Ours-1}). As seen in
table~\ref{table:accuracy}, with increased number of iterations, the
same algorithm improves overall performance (\textit{Ours-2}: 90.2\%,
\textit{Ours-4}: 91.4\%). For the stereo setup on the KITTI dataset, 3
pixels correspond to $\pm3$cm at a depth of 2 meters and $\pm80$cm at
a depth of 10m. In addition, we compare against recent
work~\cite{ramalingam2015line} on semi-dense reconstruction on the
KITTI dataset, and achieve significantly better disparity accuracy
using our approach compared to 81.2\% of~\cite{ramalingam2015line}
. In Table~\ref{table:accuracy} below, we compare the disparities
computed by our proposed method, and compare against existing stereo
matching implementations, including Semi-Global Matching, ELAS, and
Line-Sweep. We do note that the main reason for reduced accuracy
compared to state-of-the-art methods is due to the local nature of the
algorithm, as compared to the global regularization methods used in
SGBM and ELAS. We visualize the results of our proposed method in
Figure~\ref{fig:edge-disparity} with the corresponding ground truth
disparities.

\begin{table}[h]
\centering
\footnotesize
{
\setlength{\tabcolsep}{0.2em}
\begin{tabular}{lcccc}
\toprule
  \multirow{2}{*}{\textbf{Method}} & \multicolumn{4}{c}{\textbf{Accuracy (\%)}}\vspace{0.5mm}\\ 
  \multicolumn{1}{c}{} & $<2$px & $<3$px & $<4$px & $<5$px\\\midrule

SGBM~\cite{hirschmuller2005accurate}     & 89.0 & 93.9 & 95.6  & 96.5 \\ 
ELAS~\cite{geiger2011stereoscan}    & 92.7 & 96.1 & 97.3 & 97.9\\ 
Line-Sweep~\cite{ramalingam2015line}      & 72.6 & 81.2 & 84.7 & 86.7\\ 
\textit{Ours-1}$^\dagger$  &  83.1 & 89.9 & 92.9 & 94.7 \\ 
\textit{Ours-2}$^\dagger$  & 83.5 & 90.2 & 93.2 & 94.9\\ 
\textit{Ours-4}$^\dagger$  & 85.4  & 91.4 & 94.0 & 95.5 \\\bottomrule %
\end{tabular}\\\vspace{1mm}
$^\dagger$\scriptsize{The number next to
      the method indicates the number of iterations the algorithm is allowed to run.}
}
\caption{Analysis of accuracy of our system on the KITTI
dataset~\cite{Geiger2012CVPR}, as compared to popular stereo
implementations including OpenCV's Semi-Global
Block-Matching~\cite{hirschmuller2005accurate},
ELAS~\cite{geiger2011stereoscan} and
Line-Sweep~\cite{ramalingam2015line}. The number next to the method
indicates the number of iterations the algorithm is allowed to
run. The accuracy results are evaluated \textbf{only} over
high-gradient (semi-dense) regions in the image.}
\label{table:accuracy}
\vspace{-3mm}
\end{table}

\vspace{2mm}
\textbf{Stereo Reconstruction} In this section, we show the qualitative
performance of our stereo disparity estimation approach via stereo
reconstructions fused over multiple frames from a moving camera. We use the
stereo imagery from the KITTI dataset, and the corresponding ground truth poses
to reconstruct scenes over a short window time frame to qualitatively illustrate
the stereo matching consistency our approach provides. In
Figure~\ref{fig:scene-reconstructions}, we show our reconstruction results from various
sequences. The reconstructions of building facades, cars, road terrain, and road curbs
are well-detailed with little noise. Furthermore, unstructured and thin
occluding edges such
as trees, and their trunks are also well reconstructed. See video via
the following
\href{http://people.csail.mit.edu/spillai/projects/fast-stereo-reconstruction/pillai_fast_stereo16.mp4}{link}\footnote{\scriptsize\url{http://people.csail.mit.edu/spillai/projects/fast-stereo-reconstruction/pillai_fast_stereo16.mp4}}. 

\vspace{2mm}
\textbf{Runtime Performance} Most existing stereo matching algorithms
have focused their efforts on the accuracy, without much regard for
the runtime performance of these systems. In this work, we focus on
the potential benefits and trade-offs of stereo matching accuracy and
runtime performance. Due to the iterative nature of our proposed
method, we show that our approach can be tuned to various accuracy and
runtime operational levels, particularly beneficial for robotics
applications. In our experiments (Table~\ref{table:runtime}
and~\ref{table:runtime-ours}), we evaluate the runtime performance of
our proposed method across several standard image resolutions ranging
from WVGA (320x240) to HD1080 (1920x1080). For the common stereo image
resolutions (800x600), our approach provides a speed-up factor of 32x
for the single-pass stereo matching case, and a factor of 12x for the
two-pass stereo matching case, as compared to OpenCV's
SGBM~\cite{hirschmuller2005accurate} implementation.

\begin{table}[!h]
\centering
\footnotesize
\begin{tabular}{lccc}
\toprule
\textbf{Method} & \textbf{Accuracy (\%)} & \textbf{Run-time (Hz/ms)} &
                                                                 \textbf{Speed-up}\\ \midrule
SGBM~\cite{hirschmuller2005accurate}   &  93.9 & ~2.8 Hz / 351.9 ms & ~~~1x\\ 
ELAS~\cite{geiger2011stereoscan}   &  \textbf{96.1} & ~6.2 Hz / 160.9 ms & ~2.1x\\ 
Line-Sweep~\cite{ramalingam2015line} &  81.2 & 14.2 Hz / ~70.0 ms & ~~~5x\\ 
\textit{Ours-1}$^\dagger$ &  89.9 & \textbf{92.2 Hz / ~10.8 ms} & \textbf{32.4x}\\ 
\textit{Ours-2}$^\dagger$ &  90.2 & \textbf{34.6 Hz / ~28.9 ms} & \textbf{12.2x}\\ 
\textit{Ours-4}$^\dagger$ &  91.4 & \textbf{17.2 Hz / ~58.2 ms} & \textbf{~6.0x}\\
\bottomrule %
\end{tabular}\\\vspace{1mm}
\caption{Analysis of run-time performance of our system on the KITTI (1241 x 376
  px, 0.46 MP)
  dataset~\cite{Geiger2012CVPR}, as compared to popular stereo implementations
including OpenCV's Semi-Global Block-Matching~\cite{hirschmuller2005accurate}
and ELAS~\cite{geiger2011stereoscan}. The number next to the method indicates
the number of iterations the algorithm is allowed to run. We achieve comparable
performance, with a run-time speed-up of approximately \textbf{32
x}. Accuracy is reported for disparities that are within 3 pixels of ground
truth. }
\label{table:runtime}
\vspace{-3mm}
\end{table}



\begin{table}[h]
\centering
\footnotesize
{
\setlength{\tabcolsep}{0.2em}
\begin{tabular}{lcccccc}
\toprule
\multirow{2}{*}{\textbf{Method}} &  & \multicolumn{5}{c}{\textbf{Image Resolution
                                     (px)}}\vspace{0.5mm}\\ 
\multicolumn{2}{c}{} & 320x240 & 640x480 & 800x600 & 1280x720 & 1920x1080\\\midrule

SGBM~\cite{hirschmuller2005accurate}   &
\multirow{5}{*}{\rotatebox[origin=c]{90}{\textbf{Runtime (ms)}}} & 53.4 & 216.7 & 360.0 & 763.7
                                                                & 1873.7\\ 
ELAS~\cite{geiger2011stereoscan}   & &  22.7 & 107.2 & 170.3 & 332.7 & ~650.9\\ 
\textit{Ours-1}$^\dagger$ & &  \textbf{~3.0} & \textbf{~~8.2} & \textbf{~10.9} & \textbf{~18.2} & \textbf{~~35.9}\\ 
\textit{Ours-2}$^\dagger$ & &  \textbf{~6.4} & \textbf{~19.2} & \textbf{~27.4} & \textbf{~46.0} & \textbf{~~81.0}\\ 
\textit{Ours-4}$^\dagger$ & &  \textbf{18.7} &  \textbf{~64.9} &  \textbf{~99.2} & \textbf{172.9} & \textbf{~287.2}\\\bottomrule %
\end{tabular}\\\vspace{1mm}
}
\caption{\textit{Running Time vs. Image Resolution:} We compare the runtime
      performance of our proposed approach (\textit{Ours}) with existing
      state-of-the-art solutions for varied image resolutions. As shown in the table,
      our proposed stereo algorithm performs an order of magnitude faster that other
      popular approaches for high-resolution (720P) stereo imagery. The number next to
      the method indicates the number of iterations the algorithm is allowed to run.}
\label{table:runtime-ours}
\vspace{-3mm}
\end{table}




%
\begin{figure*}[t]
  \centering
  {\renewcommand{\arraystretch}{0.4} 
   {\setlength{\tabcolsep}{0.2mm}
    \begin{tabular}{ccc}
    \includegraphics[width=0.68\columnwidth]{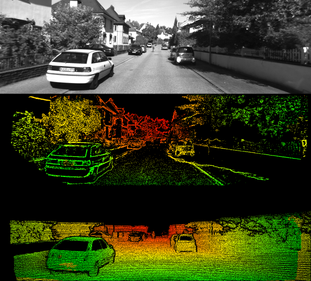}&
    \includegraphics[width=0.68\columnwidth]{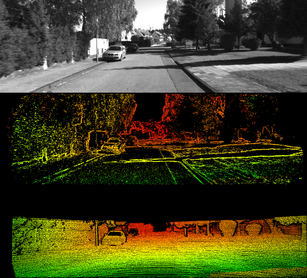}&
    \includegraphics[width=0.68\columnwidth]{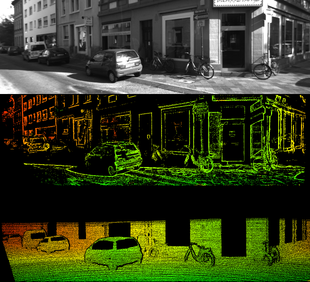}\\
    \includegraphics[width=0.68\columnwidth]{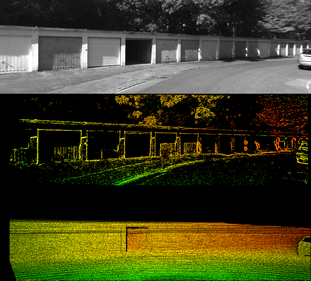}&
    \includegraphics[width=0.68\columnwidth]{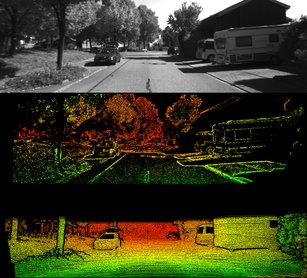}&
    \includegraphics[width=0.68\columnwidth]{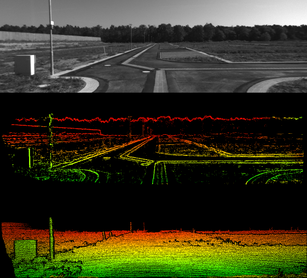}\\
  \end{tabular}}}
  \caption{Illustrations of our proposed stereo disparity estimation method
    (\textit{Ours-2}, \textbf{row 2}) on the KITTI dataset with corresponding ground truth
    estimates (\textbf{row 3}) obtained from projecting Velodyne data on to the left
    camera. Despite its short execution time, our approach shows accurate estimates
    of disparities for a variety of scenes. The ground truth estimates are provided
    as reference, and are valid points that fall below the horizon. Similar colors
    indicate similar depths at which points are registered.}
  \label{fig:edge-disparity}
\vspace{-4mm}
\end{figure*}

%
\begin{figure*}[t]
  \centering
  {\renewcommand{\arraystretch}{0.6} 
   {\setlength{\tabcolsep}{0.4mm}
    \begin{tabular}{ccc}
    \includegraphics[width=2\columnwidth]{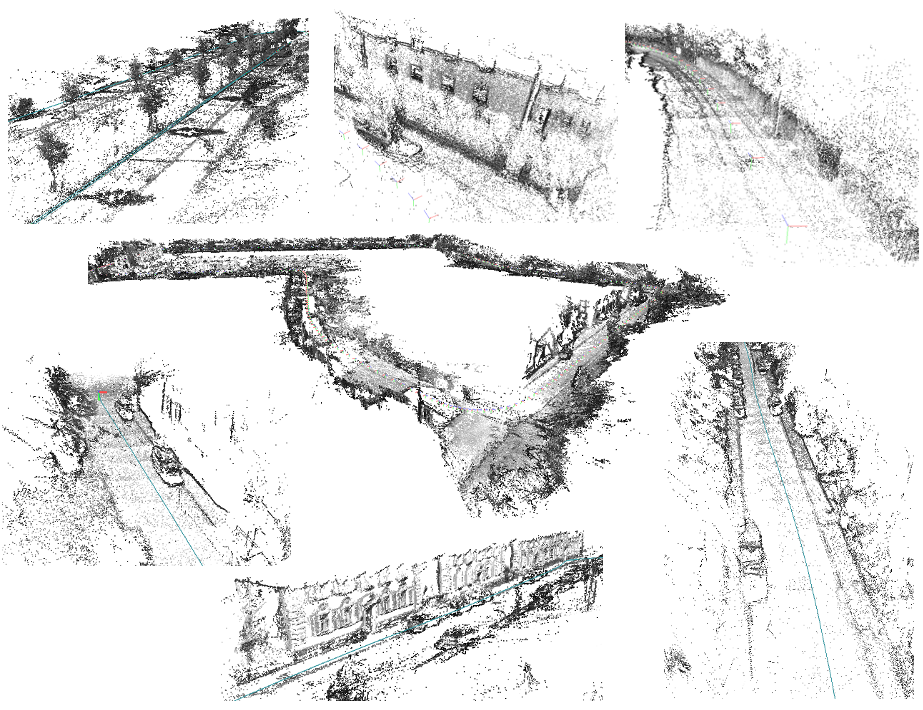}
  \end{tabular}}}
\caption{Illustrated above are various scenes reconstructed using our proposed
  stereo matching approach. We use the ground truth poses from the KITTI dataset
  to merge the reconstructions from multiple frames, qualitatively showing the
  consistency in stereo disparity estimation of our approach.}
  \label{fig:scene-reconstructions}
\vspace{-4mm}
\end{figure*}


\subsection{Evaluation on Stereo Hardware} With the advent of the USB3 standard,
high-framerate stereo cameras have now started to become mainstream. These
devices open the door to newer data throughput capacities, however, existing
state-of-the-art stereo algorithms fail to meet such high throughput
requirements. To this end, in addition to the KITTI dataset, we benchmark our
proposed method on two different stereo platforms including the BumbleBee2, and
the newly introduced USB3-driven ZED Stereo Camera.  The Bumblebee2 (12cm
baseline) operates at 48 FPS providing gray-scale stereo imagery at a resolution
of 648x488, while the ZED Camera is configured to operate at 60Hz with a
resolution of 1280x720. In our experiments, we compare the disparities estimated
from our approach against that of SGBM and report results on its accuracy and
runtime performance (see Table~\ref{table:accuracy-hardware}).

\begin{table}[h]
\centering
\footnotesize
{
\setlength{\tabcolsep}{0.4em}
\begin{tabular}{lcc}
\toprule

\multirow{2}{*}{\textbf{Method}} & \multicolumn{2}{c}{\textbf{Accuracy~(\%)}} \vspace{0.5mm}\\
                                 & \textbf{BumbleBee2} & \textbf{ZED}\\ \midrule
ELAS~\cite{geiger2011stereoscan}   & 81.1  & \textbf{91.6} \\ 
Line-Sweep~\cite{ramalingam2015line} & 83.9  & 77.2 \\ 
\textit{Ours-1}$^\dagger$ & 89.6  & 87.5  \\ 
\textit{Ours-2}$^\dagger$ & \textbf{90.8}  & 87.3  \\ 
\bottomrule %
\end{tabular}\\\vspace{1mm}
}
\caption{Analysis of accuracy of our system on the BumbleBee2 and ZED Stereo
  Camera, with Semi-Global Block-Matching~\cite{hirschmuller2005accurate}
  considered as ground truth. We compare against other stereo implementations
including ELAS and Line-Sweep and report the accuracy for disparities that are
within 3 pixels of ground truth. The number next to the method indicates
the number of iterations the algorithm is allowed to run. }
\label{table:accuracy-hardware}

\end{table}


\subsection{Implementation} We use the high-speed sparse-stereo
implementation of~\cite{schauwecker2012new}, and the Delaunay
Tessellation is performed via the
Triangle\footnote{\scriptsize\url{https://www.cs.cmu.edu/~quake/triangle.html}}
library for the initial set of support tessellation. Besides the 5x5
Census-based block matching that is implemented using specialized SSE
instructions~\cite{schauwecker2012new}, the rest of the code is
implemented on a single-CPU thread in C++, without any specialized
instruction sets or GPU-specific code. All the results of our code are
tested on an Intel(R) Core(TM) i7-3920XM CPU @ 2.90GHz. We do note
that while our current implementation refines disparities every
iteration in batch, this step can be highly-parallel and asynchronous
due to the recursive nature of the refinement over the tessellated
structure.
\section{Discussion} Several robotics applications adhere to strict
computational budgets and runtime requirements, depending on their
task domain. Some systems require the need to actively adapt to
varying design requirements and conditions, and adjust parameters
accordingly. In the context of mapping and navigation, robots may need
to map the world around them, in a slow but accurate manner, while
also requiring the ability to avoid dynamic obstacles quickly and
robustly. Such systems require the ability to dynamically change the
accuracy requirements in order to achieve their desired runtime
performance, given a fixed compute budget; this work is an attempt to
provide such capability.

Another potential application of this approach could be to generate
rapid and high-fidelity reconstructions, given a sufficiently coarse
trajectory plan or foveation. Given a reasonable
exploration-exploitation strategy, our approach can provide promising
flexibility in exploiting accurate and rich scene information, while
also being able to adjust itself to rapidly handle dynamic scenes
during the exploration stage.




\section{Conclusion} Most existing stereo matching methods have been designed to
ensure high accuracy guarantees for disparity estimation, however, sacrifice
their runtime performance as a result. In this work, we propose a novel and
high-performance, iterative and semi-dense stereo matching method, capable of running
at a peak framerate of 120Hz, with comparable accuracies to existing and popular
stereo matching solutions. By maintaining a piece-wise planar assumption, we
develop a stereo matching strategy that recursively tessellates the scene into
piece-wise planar regions so that it appropriately reconstructs it, given a
fixed runtime requirement as provided by the user. By evaluating the matching
costs for candidate planes, our approach quickly identifies planar regions, and
repeats the process for non-planar regions by introducing more stereo matches
within these regions and re-tessellating them. We compare against stereo
matching algorithms that are commonly used in robotics applications and provide
promising results of the trade-offs between matching accuracy and run-times
achievable by our proposed method, across varied stereo dataset and hardware
setups.


\balance
\bibliographystyle{templates/icra/bst/IEEEtran}
\bibliography{tex/references}



\end{document}